\def\year{2020}\relax
\documentclass[letterpaper]{article} 
\usepackage{aaai20}  
\usepackage{times}  
\usepackage{helvet} 
\usepackage{courier}  
\usepackage[hyphens]{url}  
\usepackage{graphicx} 

\usepackage{latexsym}
\usepackage{algorithm}
\usepackage{algpseudocode}
\usepackage{amsmath}
\usepackage{subfig}
\usepackage{tabularx}
\usepackage{amssymb}
\usepackage{booktabs}
\usepackage{multirow}
\usepackage{verbatim}
\usepackage{subfig}
\usepackage{tabularx}

\newcommand \newcite[1]{{\citeauthor{#1}~\shortcite{#1} }}
\newcommand \method{GRET }
\usepackage{graphicx}  
\title{GRET: Global Representation Enhanced Transformer}
\author{
\Large \textbf{Rongxiang Weng\textsuperscript{\rm 1,2}, Haoran Wei\textsuperscript{\rm 2}, Shujian Huang\textsuperscript{\rm 1}\thanks{Corresponding author}, Heng Yu\textsuperscript{\rm 2},}
\\ \Large \textbf{ Lidong Bing\textsuperscript{\rm 2}, Weihua Luo\textsuperscript{\rm 2}, Jiajun Chen\textsuperscript{\rm 1}}\\ 
\textsuperscript{\rm 1}National Key Laboratory for Novel Software Technology, Nanjing University, Nanjing, China\\ 
\textsuperscript{\rm 2}Machine Intelligence Technology Lab, Alibaba Group, Hangzhou, China \\ 
\{wengrx,funan.whr\}@alibaba-inc.com,huangsj@nju.edu.cn,
\\
\{yuheng.yh,l.bing,weihua.luowh\}@alibaba-inc.com,
chenjj@nju.edu.cn 
}
 
\begin{document}

\maketitle

\begin{abstract}
Transformer, based on the encoder-decoder framework, has achieved state-of-the-art performance on several natural language generation tasks.
The encoder maps the words in the input sentence into a sequence of hidden states, which are then fed into the decoder to generate the output sentence. 
These hidden states usually correspond to the input words and focus on capturing local information.
However, the global (sentence level) information is seldom explored, leaving room for the improvement of generation quality.
In this paper, we propose a novel global representation enhanced Transformer (GRET) to explicitly model global representation in the Transformer network. Specifically, in the proposed model, an external state is generated for the global representation from the encoder. 
The global representation is then fused into the decoder during the decoding process to improve generation quality. We conduct experiments in two text generation tasks: machine translation and text summarization. Experimental results on four WMT machine translation tasks and LCSTS text summarization task demonstrate the effectiveness of the proposed approach on natural language generation.
\end{abstract}
    
\section{Introduction}
Transformer~\cite{vaswani2017attention} has outperformed other methods on several neural language generation (NLG) tasks, like machine translation~\cite{deng2018alibaba}, text summarization~\cite{chang2018hybrid}, \textit{etc}. 
Generally, Transformer is based on the \textit{encoder-decoder framework} which consists of two modules: an encoder network and a decoder network. The encoder encodes the input sentence into a sequence of hidden states, each of which corresponds to a specific word in the sentence. The decoder generates the output sentence word by word. At each decoding time-step, the decoder performs attentive read~\cite{Luong2015Effective,vaswani2017attention} to fetch the input hidden states and decides which word to generate.
    
As mentioned above, the decoding process of Transformer only relies on the representations contained in these hidden states. However, there is evidence showing that hidden states from the encoder in Transformer only contain local representations which focus on word level information. 
For example, previous work~\cite{vaswani2017attention,devlin2018bert,song2020alignment} showed that these hidden states pay much attention to the word-to-word mapping; and the weights of attention mechanism, determining which target word will be generated, is similar to word alignment.  

As \newcite{frazier1987sentence} pointed, the global information, which is about the whole sentence in contrast to individual words, should be involved in the process of generating a sentence. Representation of such global information plays an import role in neural text generation tasks. 
In the recurrent neural network (RNN) based models~\cite{Bahdanau2015Neural}, \newcite{Chen2018} showed on text summarization task that introducing representations about global information could improve quality and reduce repetition. \newcite{lin2018de} showed on machine translation that the structure of the translated sentence will be more correct when introducing global information. These previous work shows global information is useful in current neural network based model. However, different from RNN~\cite{sutskever2014sequence,Cho2014Learning,Bahdanau2015Neural} or CNN~\cite{gehring2016convolutional,gehring2017convolutional}, although self-attention mechanism can achieve long distance dependence, there is no explicit mechanism in the Transformer to model the global representation of the whole sentence. Therefore, it is an appealing challenge to provide Transformer with such a kind of global representation.

In this paper, we divide this challenge into two issues that need to be addressed:    
1). \textit{how to model the global contextual information?} and 2). \textit{how to use global information in the generation process?}, and propose a novel global representation enhanced Transformer (GRET) to solve them.
For the first issue, we propose to generate the global representation based on local word level representations by two complementary methods in the encoding stage. 
On one hand, we adopt a modified \textit{capsule network}~\cite{sabour2017dynamic} to generate the global representation based the features extracted from local word level representations.
The local representations are generally related to the word-to-word mapping, which may be redundant or noisy. Using them to generate the global representation directly without any filtering is inadvisable.
Capsule network, which has a strong ability of feature extraction~\cite{zhao2018investigating}, can help to extract more suitable features from local states. Comparing with other networks, like CNN~\cite{krizhevsky2012imagenet}, it can see all local states at one time, and extract feature vectors after several times of deliberation.
    
On the other hand, we propose a \textit{layer-wise recurrent structure} to further strengthen the global representation. Previous work shows the representations from each layer have different aspects of meaning~\cite{peters2018deep,dou2018exploiting}, e.g. lower layer contains more syntactic information, while higher layer contains more semantic information. A complete global context should have different aspects of information. 
However, the global representation generated by the capsule network only obtain intra-layer information. 
The proposed layer-wise recurrent structure is a helpful supplement to combine inter-layer information by aggregating representations from all layers. 
These two methods can model global representation by fully utilizing different grained information from local representations.

For the second issue, we propose to use \textit{a context gating mechanism} to dynamically control how much information from the global representation should be fused into the decoder at each step.
In the generation process, every decoder states should obtain global contextual information before outputting words. And the demand from them for global information varies from word to word in the output sentence. The proposed gating mechanism could utilize the global representation effectively to improve generation quality by providing a customized representation for each state.
    
Experimental results on four WMT translation tasks, and LCSTS text summarization task show that our \method model brings significant improvements over a strong baseline and several previous researches.

\section{Approach}
Our \method model includes two steps: modeling the global representation in the encoding stage and incorporating it into the decoding process. We will describe our approach in this section based on Transformer~\cite{vaswani2017attention}.
\subsection{Modeling Global Representation}
In the encoding stage, we propose two methods for modeling the global representation at different granularity.
We firstly use capsule network to extract features from local word level representations, and generate global representation based on these features. 
Then, a layer-wise recurrent structure is adopted subsequently to strengthen the global representation by aggregating the representations from all layers of the encoder. The first method focuses on utilizing word level information to generate a sentence level representation, while the second method focuses on combining different aspects of sentence level information to obtain a more complete global representation.
    
\paragraph{Intra-layer Representation Generation} We propose to use \textit{capsules with dynamic routing} to extract the specific and suitable features from the local representations for stronger global representation modeling, which is an effective and strong feature extraction method \cite{sabour2017dynamic,zhang2018sentence}\footnote{Other details of the Capsule Network are shown in \newcite{sabour2017dynamic}.}.
Features from hidden states of the encoder are summarized into several capsules, and the weights (routes) between hidden states and capsules are updated by dynamic routing algorithm iteratively.

\begin{algorithm}[tb]
    \caption{Dynamic Routing Algorithm}
    \label{alg:dynamic routing}
    \begin{algorithmic}[1] 
    \State \textbf{procedure}: \textsc{Routing}($\textbf{H}$, $r$)
    \For{$i$ in input layer and $k$ in output layer }
    \State $b_{ki}\leftarrow0$;
    \EndFor
    \For{$r$ iterations} 
        \For{$k$ in output layer}
        \State $\textbf{c}_{k}\leftarrow\text{softmax}(\textbf{b}_{k})$;
        \EndFor
        \For{$k$ in output layer}
        \State $\textbf{u}_{k}\leftarrow q(\sum_{i}^{I}c_{ki}\textbf{h}_{i})$; 
        \\ \Comment{$\textbf{H}=\{\textbf{h}_{1},\cdots,\textbf{h}_{i},\cdots\}$}
        \EndFor
        \For{$i$ in input layer and $k$ in output layer}
        \State $b_{ki}\leftarrow b_{ki}+\textbf{h}_{i}\cdot\textbf{u}_{k}$;
        \EndFor 
    \EndFor 
    \State \textbf{return} $\textbf{U}$; 
    \Comment{$\textbf{U}=\{\textbf{u}_{1},\cdots,\textbf{u}_{k},\cdots\}$}
    \end{algorithmic}
\end{algorithm}

Formally, given an encoder of the Transformer which has $M$ layers and an input sentence $\textbf{x}=\{x_{1}, \cdots, x_{i}, \cdots, x_{I}\}$ which has $I$ words.
The sequence of hidden states $\textbf{H}^{m}=\{\textbf{h}^{m}
_{1},\cdots,\textbf{h}^{m}_{i},\cdots,\textbf{h}^{m}_{I}\}$ from the $m^\text{th}$ layer of the encoder is computed by
\begin{align}
    \textbf{H}^{m}=\text{LN}(\text{SAN}(\textbf{Q}^{m}_{e},\textbf{K}_{e}^{m-1},\textbf{V}_{e}^{m-1})),
\end{align} 
where the $\textbf{Q}^{m}_{e}$, $\textbf{K}^{m-1}_{e}$ and $\textbf{V}^{m-1}_{e}$ are query, key and value vectors which are same as $\textbf{H}^{m-1}$, the hidden states from the $m-1^\text{th}$ layer. The $\text{LN}(\cdot)$ and $\text{SAN}(\cdot)$ are layer normalization function~\cite{ba2016layer} and self-attention network~\cite{vaswani2017attention}, respectively. We omit the residual network here.
    
Then, the capsules $\textbf{U}^{m}$ with size of $K$ are generated by $\textbf{H}^{m}$. Specifically, the $k^\text{th}$ capsule $\textbf{u}^m_{k}$ is computed by 
\begin{align} 
    \textbf{u}^m_{k}&=q(\sum_{i}^{I}c_{ki}\hat{\textbf{h}}^{m}_{i}),~c_{ki} \in \textbf{c}_{k},  \\
    \hat{\textbf{h}}^{m}_{i}& = \textbf{W}_{k}\textbf{h}^{m}_{i},
\end{align}
where $q(\cdot)$ is non-linear squash function~\cite{sabour2017dynamic}:
\begin{align}
    \text{squash}(\textbf{t})=\frac{||\textbf{t}||^{2}}{1+||\textbf{t}||^{2}}\frac{\textbf{t}}{||\textbf{t}||},
\end{align}
and $\textbf{c}_{k} $ is computed by
\begin{align}
    \textbf{c}_{k} = \text{softmax}(\textbf{b}_{k}),~\textbf{b}_{k} \in \textbf{B},
\end{align}
where the matrix $\textbf{B}$ is initialized by zero and whose row and column are $K$ and $I$, respectively. This matrix will be updated when all capsules are produced.
\begin{align}
    \textbf{B} = \textbf{B} + \textbf{U}^{m\top} \cdot \textbf{H}^m.
\end{align}
The algorithm is shown in Algorithm \ref{alg:dynamic routing}. 
The sequence of capsules $\textbf{U}^{m}$ could be used to generate the global representation.

\begin{figure}[t]
    \centering
    \includegraphics[scale = 0.40]{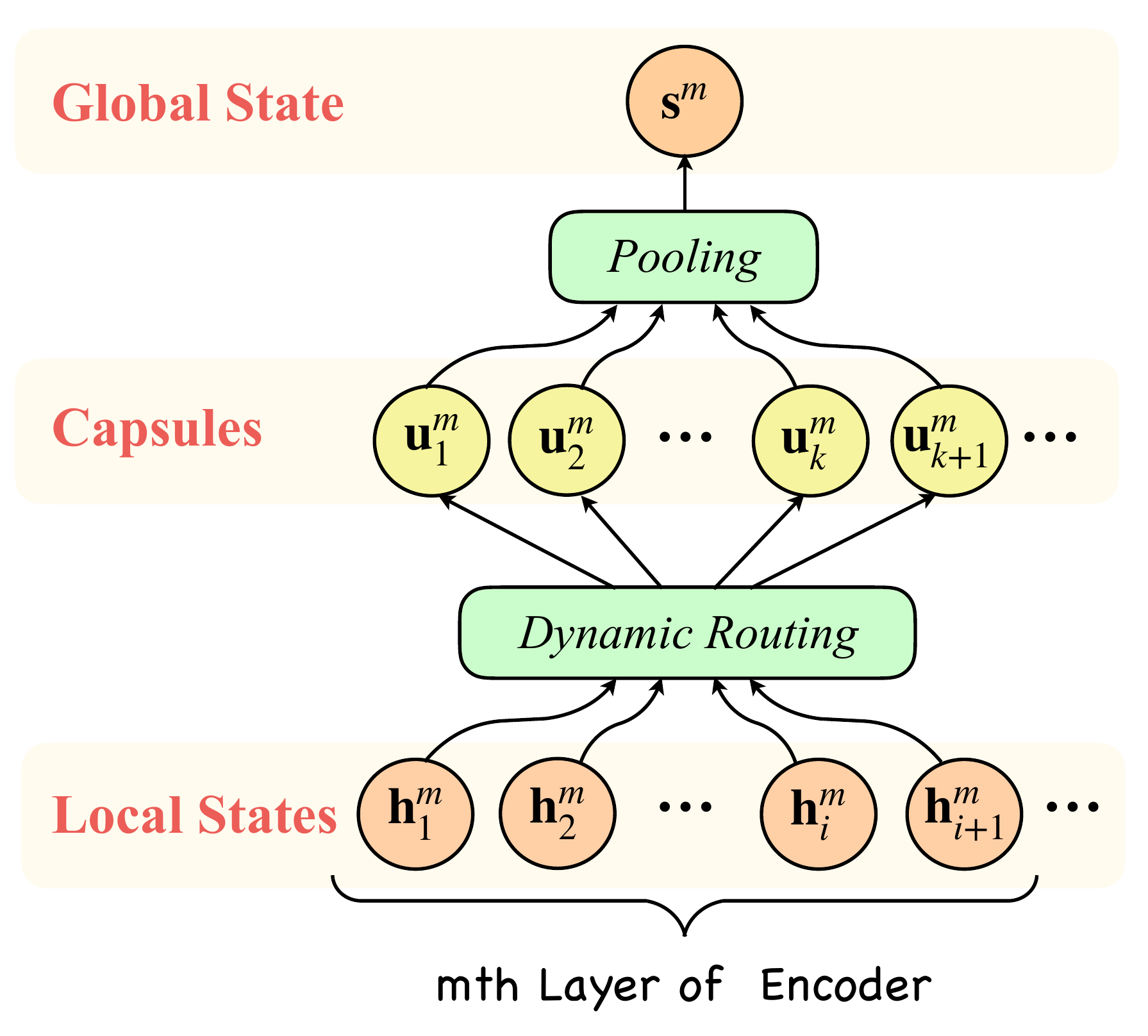}
    \caption{\label{fig: sentence representation} The overview of generating the global representation with capsule network.}
\end{figure}

Different from the original capsules network which use a concatenation method to generate the final representation, we use an \textit{attentive pooling method} to generate the global representation\footnote{Typically, the concatenation and other pooling methods, e.g. mean pooling, could be used here easily, but they will decrease 0.1$\sim$0.2 BLEU in machine translation experiment.}. Formally, in the $m^\text{th}$ layer, the global representation is computed by
\begin{align}
\textbf{s}^{m}&=\text{FFN}(\sum_{k=1}^{K}a_{k}\textbf{u}^{m}_{k}),
\label{eq: ffn} \\
a_{k}&=\frac{\text{exp}(\hat{\textbf{s}}^{m} \cdot \textbf{u}^{m}_{k})}{\sum_{t=1}^{K}\text{exp}(\hat{\textbf{s}}^{m} \cdot \textbf{u}^{m}_{t})},
\end{align}
where $\text{FFN}(\cdot)$ is a feed-forward network and the $\hat{\textbf{s}}^{m}$ is computed by
\begin{align}
    \textbf{s}^{m}=\text{FFN}(\frac{1}{K}\sum^{K}_{k=1}{\textbf{u}^{m}_{k}}). \label{eq: average}
\end{align}
    
This \textit{attentive} method can consider the different roles of the capsules and better model the global representation. The overview of the process of generating the global representation are shown in Figure~\ref{fig: sentence representation}.

\paragraph{Inter-layer Representation Aggregation} Traditionally, the Transformer model only fed the last layer's hidden states $\textbf{H}^{M}$ as representations of input sentence to the decoder to generate the output sentence. Following this, we can feed the last layer's global representation $\textbf{s}^{M}$ into the decoder directly. However, current global representation only contain the intra-layer information, the other layers' representations are ignored, which were shown to have different aspects of meaning in previous work~\cite{wang2018multi,dou2018exploiting}. 
Based on this intuition, we propose a \textit{layer-wise recurrent structure} to aggregate the representations generated by employing the capsule network on all layers of the encoder to model a complete global representation.

\begin{figure}
    \centering
    \includegraphics[scale = 0.40]{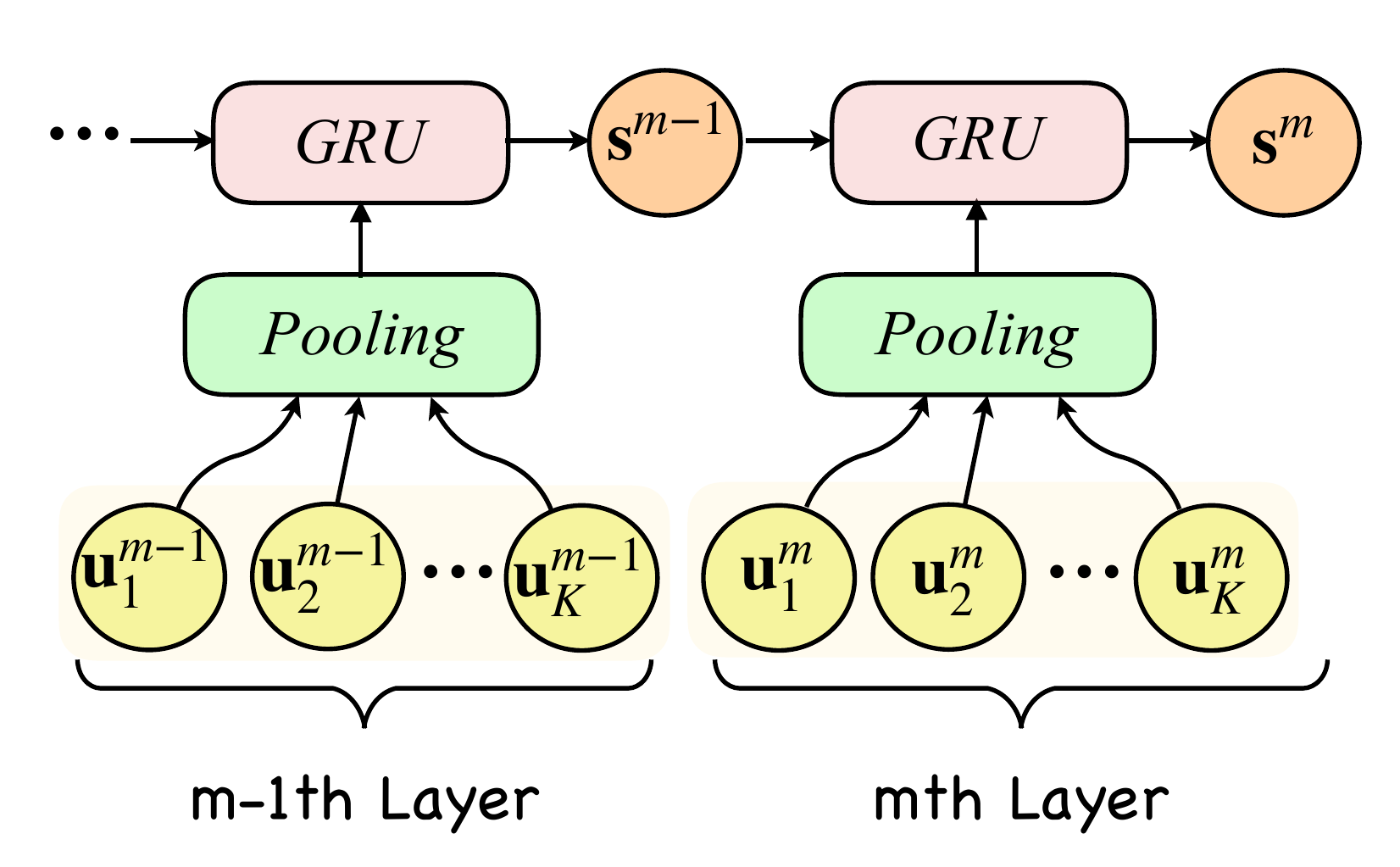}
    \caption{\label{fig: deep} The overview of the layer-wise recurrent structure.}
\end{figure}

The layer-wise recurrent structure aggregates each layer's intra global state by a gated recurrent unit~\cite[GRU]{Cho2014Learning} which could achieve different aspects of information from the previous layer's global representation.
Formally, we adjust the computing method of $\textbf{s}^{m}$ by
\begin{align}
    \textbf{s}^{m}=\text{GRU}(\text{ATP}(\textbf{U}^{m}),\textbf{s}^{m-1}),
\end{align}
where the $\text{ATP}(\cdot)$  is the attentive pooling function computed by Eq \ref{eq: ffn}-\ref{eq: average}.
The GRU unit can control the information flow by forgetting useless information and capturing suitable information, which can aggregate previous layer's representations usefully.
The layer-wise recurrent structure could achieve a more exquisite and complete representation. Moreover, the proposed structure only need one more step in the encoding stage which is not time-consuming. 
The overview of the aggregation structure is shown in Figure \ref{fig: deep}. 

\subsection{Incorporating into the Decoding Process}
Before generating the output word, each decoder state should consider the global contextual information. We combine the global representation in decoding process with an additive operation to the last layer of the decoder guiding the states output true words. However, the demand for the global information of each target word is different. Thus, we propose \textit{a context gating mechanism} which can provide specific information according to each decoder hidden state. 
    
\begin{figure}[t]
    \centering
    \includegraphics[scale = 0.40]{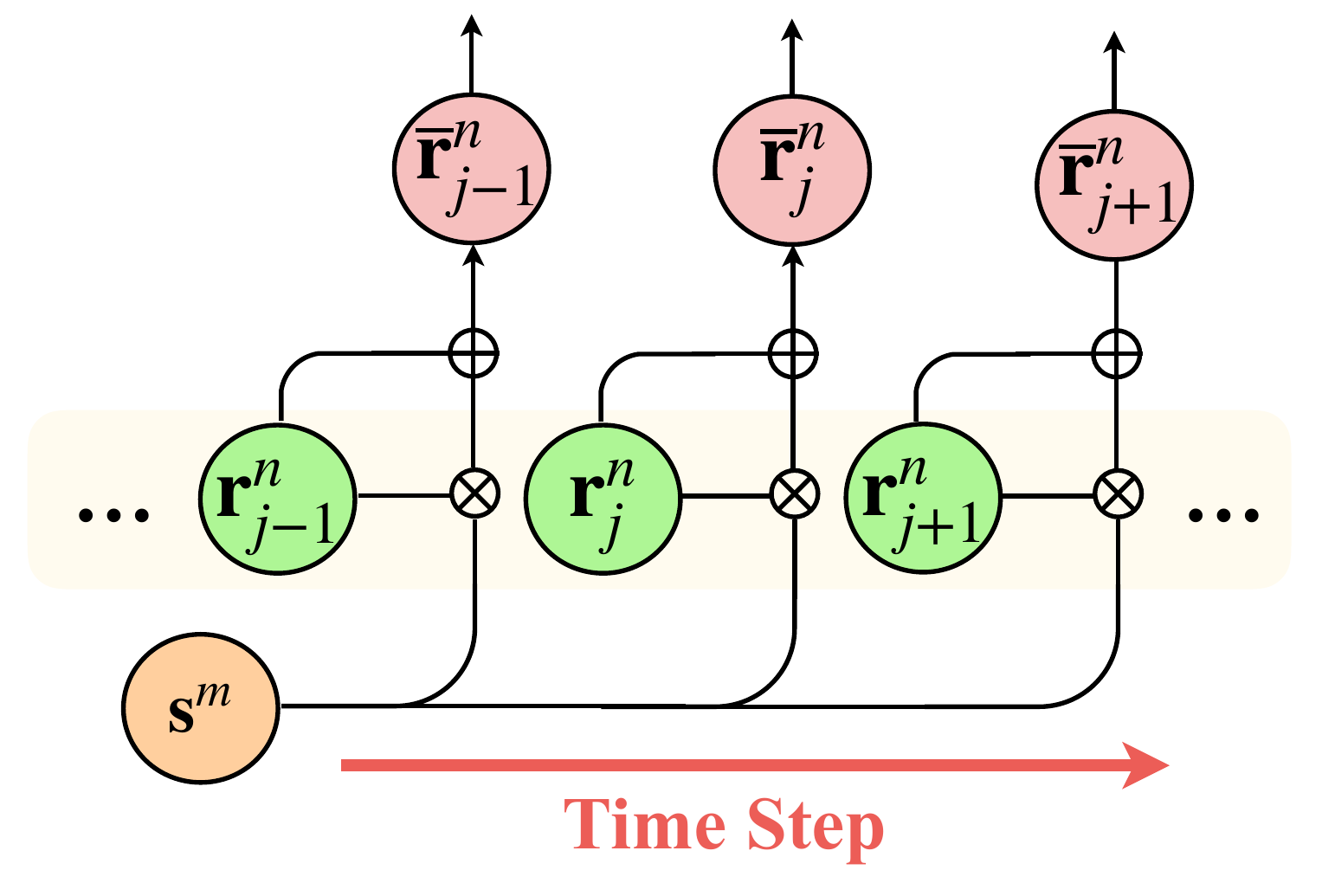}
    \caption{\label{fig: dg} The context gating mechanism of fusing the global representation into decoding stage.}
\end{figure}

Specifically, given an decoder which has $N$ layers and the target sentence $\textbf{y}$ which has $J$ words in the training stage, the hidden states $\textbf{R}^{N}=\{\textbf{r}^{N}_{1},\cdots,\textbf{r}^{N}_{j},\cdots,\textbf{r}^{N}_{J}\}$ from the $N^\text{th}$ layer of the decoder is computed by 
\begin{align}
    \textbf{R}^{N}=\text{LN}(&\text{SAN}(\textbf{Q}^{N}_{d},\textbf{K}_{d}^{N-1},\textbf{V}_{d}^{N-1}) \nonumber
    \\&+\text{SAN}(\textbf{Q}^{N}_{d},\textbf{K}_{e}^{M},\textbf{V}_{e}^{M})),
\end{align} 
where $\textbf{Q}^{N}_{d}$, $\textbf{K}^{N-1}_{d}$ and $\textbf{V}^{N-1}_{d}$ are hidden states $\textbf{R}^{N-1}$ from $N-1^{\text{th}}$ layer. The $\textbf{K}_{e}^{M}$ and $\textbf{V}_{e}^{M}$ are same as $\textbf{H}^{M}$. We omit the residual network here.

For each hidden state $\textbf{r}^{N}_{j}$ from $\textbf{R}^{N}$, the context gate is calculated by:
\begin{align}
\textbf{g}_{j}=\text{sigmoid}(\textbf{r}^{N}_{j},\textbf{s}^{M}).
\end{align}
The new state, which contains the needed global information, is computed by:
\begin{align}
\overline{\textbf{r}}^{N}_{j} = \textbf{r}^{N}_{j} + \textbf{s}^{M}_{j}*\textbf{g}.
\end{align}
Then, the output probability is calculated by the output layer's hidden state:
\begin{align}
P(y_{j}|y_{<j},\textbf{x})&=\text{softmax}(\text{FFN}(\overline{\textbf{r}}^{N}_{j})). \label{prob}
\end{align}
This method enables each state to achieve it's customized global information. The overview is shown in Figure \ref{fig: dg}.

\subsection{Training}
The training process of our \method model is same as the standard Transformer. The networks is optimized by maximizing the likelihood of the output sentence $\textbf{y}$ given input sentence $\textbf{x}$, denoted by $\mathcal{L}_\text{trans}$.
\begin{align}
\mathcal{L}_{\text{trans}}=\frac{1}{J}\sum_{j=1}^{J}\log P(y_{j}|y_{<j},\textbf{x}),
\end{align}
where $P(y_{j}|y_{<j},\textbf{x})$ is defined in Equation~\ref{prob}.

\section{Experiment}

\subsection{Implementation Detail}
\paragraph{Data-sets} We conduct experiments on machine translation and text summarization tasks. In machine translation, we employ our approach on four language pairs: Chinese to English (ZH$\rightarrow$EN), English to German (EN$\rightarrow$DE), German to English (DE$\rightarrow$EN), and Romanian to English (RO$\rightarrow$EN)~\footnote{http://www.statmt.org/wmt17/translation-task.html}. In text summarization, we use LCSTS~\cite{hu2015lcsts}~\footnote{http://icrc.hitsz.edu.cn/Article/show/139.html} to evaluate the proposed method. These data-sets are public and widely used in previous work, which will make other researchers replicate our work easily. 

In machine translation, on the ZH$\rightarrow$EN task, we use WMT17 as training set which consists of about 7.5M sentence pairs. We use {\texttt{newsdev2017}} as validation set and \texttt{newstest2017} as test set which have 2002 and 2001 sentence pairs, respectively.
On the EN$\rightarrow$DE and DE$\rightarrow$EN tasks, we use WMT14 as training set which consists of about 4.5M sentence pairs. We use {\texttt{newstest2013}} as validation set and \texttt{newstest2014} as test set which have 2169 and 3000 sentence pairs, respectively.
On the RO$\rightarrow$EN task, we use WMT16 as training set which consists of about 0.6M sentence pairs. We use {\texttt{newstest2015}} as validation set and \texttt{newstest2016} as test set which has 3000 and 3002 sentence pairs, respectively.

In text summarization, following in \newcite{hu2015lcsts}, we use PART I as training set which consists of 2M sentence pairs. We use the subsets of PART II and PART III scored from 3 to 5 as validation and test sets which consists of 8685 and 725 sentence pairs, respectively.  
    
\begin{table*}[ht]\footnotesize
    \centering
    \begin{tabular}{l||c|c|c|c}
        \toprule
        Model & ZH$\rightarrow$EN&EN$\rightarrow$DE&DE$\rightarrow$EN&RO$\rightarrow$EN  \\
        \hline
        $\text{Transformer}^{*}$~\cite{vaswani2017attention}&$-$&27.3&$-$&$-$\\
        $\text{Transformer}^{*}$~\cite{hassan2018achieving}& 24.13&$-$&$-$&$-$\\
        $\text{Transformer}^{*}$~\cite{gu2017non}&$-$&27.02&$-$&31.76 \\
        \hline
        $\text{DeepRepre}^{*}$~\cite{dou2018exploiting} & 24.76&28.78&$-$&$-$ \\
        $\text{Localness}^{*}$~\cite{yang2018modeling} &24.96&28.54&$-$&$-$ \\
        $\text{RelPos}^{*}$~\cite{shaw2018self} & 24.53&27.94&$-$&$-$ \\
        $\text{Context-aware}^{*}$~\cite{yang2019context}& 24.67&28.26&$-$&$-$ \\
        $\text{GDR}^{*}$~\cite{zheng2019dynamic}&$-$&28.10&$-$&$-$\\
        \hline
        Transformer &24.31&27.20&32.34&32.17 \\
        \method & 25.53$^{\ddagger}$&28.46$^{\dagger}$&33.79$^{\ddagger}$&33.06$^{\ddagger}$ \\
        \bottomrule
    \end{tabular}
    \caption{The comparison of our \method, Transformer baseline and related work on the WMT17 Chinese to English (ZH$\rightarrow$EN), WMT14 English to German (EN$\rightarrow$DE) and German to English (DE$\rightarrow$EN), and WMT16 Romania to English (RO$\rightarrow$EN) tasks (* indicates the results came from their paper, $\dagger/\ddagger$ indicate significantly better than the baseline ($p<0.05/0.01$)).}
    \label{tab:mt_r}
\end{table*}

\begin{table*}[t]\footnotesize
    \centering
    \begin{tabular}{l||c|c|c}
        \toprule
        Model & ROUGE-1&ROUGE-2&ROUGE-L \\
        \hline
        $\text{RNNSearch}^{*}$~\cite{hu2015lcsts}&30.79&$-$&$-$ \\
        $\text{CopyNet}^{*}$~\cite{gu2016incorporating}&34.4&21.6&31.3 \\
        $\text{MRT}^{*}$~\cite{ayana2016neural} & 37.87&25.43&35.33  \\
        $\text{AC-ABS}^{*}$~\cite{li2018actor} & 37.51&24.68&35.02  \\
        $\text{CGU}^{*}$~\cite{Lin2018GlobalEF}&39.4&26.9&36.5\\
        $\text{Transformer}^{*}$~\cite{chang2018hybrid}& 42.35 & 29.38&39.23\\
        \hline
        Transformer &43.14&29.26&39.72\\
        \method & 44.77 & 30.96&41.21 \\
        \bottomrule
    \end{tabular}
    \caption{The comparison of our \method, Transformer baseline and related work on the LCSTS text summarization task (* indicates the results came from their paper).}
    \label{tab: lcsts}
\end{table*} 

\paragraph{Settings} In machine translation, we apply byte pair encoding (BPE)~\cite{sennrich2015neural} to all language pairs and limit the vocabulary size to 32K. In text summarization, we limit the vocabulary size to 3500 based on the character level. Out-of-vocabulary words and chars are replaced by the special token \emph{UNK}. 

For the Transformer, we set the dimension of the input and output of all layers as 512, and that of the feed-forward layer to 2048. 
We employ 8 parallel attention heads. The number of layers for the encoder and decoder are 6. 
Sentence pairs are batched together by approximate sentence length. Each batch has 50 sentence and the maximum length of a sentence is limited to 100. 
We set the value of dropout rate to 0.1. We use the Adam~\cite{kingma2014adam} to update the parameters, and the learning rate was varied under a warm-up strategy with 4000 steps~\cite{vaswani2017attention}. Other details are shown in \newcite{vaswani2017attention}. The number of capsules is set 32 and the default time of iteration is set 3. The training time of the Transformer is about 6 days on the DE$\rightarrow$EN task. And the training time of the \method model is about 12 hours when using the parameters of baseline as initialization.

After the training stage, we use beam search for heuristic decoding, and the beam size is set to 4. 
We measure translation quality with the NIST-BLEU~\cite{Papineni2002bleu} and summarization quality with the ROUGE~\cite{lin2004rouge}. 

\subsection{Main Results}
\paragraph{Machine Translation} We employ the proposed \method model on four machine translation tasks. All results are summarized in Table \ref{tab:mt_r}.
For fair comparison, we reported several Transformer baselines with same settings reported by previous work~\cite{vaswani2017attention,hassan2018achieving,gu2017non} and researches about enhancing local word level representations~\cite{dou2018exploiting,yang2018modeling,shaw2018self,yang2019context}.

The results on the WMT17 ZH$\rightarrow$EN task are shown in the second column of Table \ref{tab:mt_r}.  
The improvement of our \method model could be up to 1.22 based on a strong baseline system, which outperforms all previous work we reported. 
To our best knowledge, our approach attains the state-of-the-art in relevant researches.

Then, the results on the WMT14 EN$\rightarrow$DE and DE$\rightarrow$EN tasks, which is the most widely used data-set recently, are shown in the third and fourth columns. 
The \method model could attain 28.46 BLEU (+1.26) on the EN$\rightarrow$DE and 33.79 BLEU (+1.45) on the DE$\rightarrow$EN, which are competitive results compared with previous studies.

To verify the generality of our approach, we also experiment it on low resource language pair of the WMT16 RO$\rightarrow$EN task. Results are shown in the last column. 
The improvement of the \method is 0.89 BLEU, which is a material improvement in low resource language pair. And it shows that proposed methods could improve translation quality in low resource scenario.

Experimental results on four machine translation tasks show that modeling global representation in the current Transformer network is a general approach, which is not limited by the language or size of training data, for improving translation quality.

\begin{table*}[t]\footnotesize
    \centering
    \begin{tabular}{l|ccc||cccc}
        \toprule
        Model&Capsule&Aggregate&Gate&\#Param&Inference&BLEU&$\Delta$\\
        \hline
        Transformer & $-$& $-$& $-$& 61.9M& 1.00x &27.20 & $-$\\
        \hline
        \multirow{8}{*}{\textit{Our Approach}}& & & & 61.9M& 0.99x&27.39&+0.19\\
        &\checkmark & & & 63.6M& 0.87x&28.02&+0.82\\
        &\checkmark &\checkmark & & 68.1M& 0.82x&28.32&+1.02\\
        &\checkmark & &\checkmark & 63.6M& 0.86x&28.23&+1.03\\
        &&\checkmark  & & 66.6M& 0.95x&27.81&+0.61\\
        &&\checkmark  &\checkmark & 66.8M& 0.93x&27.76&+0.56\\
        &&&\checkmark   & 62.1M& 0.98x&27.53&+0.33\\
        &\checkmark &\checkmark &\checkmark & 68.3M& 0.81x&28.46&+1.26\\

        \bottomrule
        
    \end{tabular}
    \caption{Ablation study on the WMT14 English to German (EN$\rightarrow$DE) machine translation task.}
    \label{tab: ablation}
\end{table*}

\paragraph{Text Summarization} Besides machine translation, we also employ proposed methods in text summarization, a monolingual generation task, which is an important and typical task in natural language generation. 

The results are shown in Table \ref{tab: lcsts}, we also reports several popular methods in this data-set as a comparison. Our approach achieves considerable improvements in ROUGE-1/2/L (+1.63/+1.70/+1.49) and outperforms other work with same settings. The improvement on text summarization is even more than machine translation. Compared with machine translation, text summarization focuses more on extracting suitable information from the input sentence, which is an advantage of the \method model.

Experiments on the two tasks also show that our approach could work on different types of language generation task and may improve the performance of other text generation tasks.

\begin{table}[t]\footnotesize
    \centering
    \begin{tabular}{l||c|c|c}
    \toprule
    Model &\#Param&Inference&BLEU \\
    \hline
    Transformer-Base  &61.9M&1.00x&27.20 \\
    
    \hline
    GTR-Base  &68.3M&0.81x&28.46\\
    \hline
    \hline
    Transformer-Big  &249M&0.59x&28.47 \\
    
    \hline
    \textsc{GReT}-Big  &273M&0.56x&29.33\\
    \bottomrule
    \end{tabular}
    \caption{The comparison of \method and Transformer with \textit{big} setting~\cite{vaswani2017attention} on the EN$\rightarrow$DE task.}
    \label{tab: bigsize}
\end{table}

\begin{figure}[t]
    \centering
    \includegraphics[scale = 0.25]{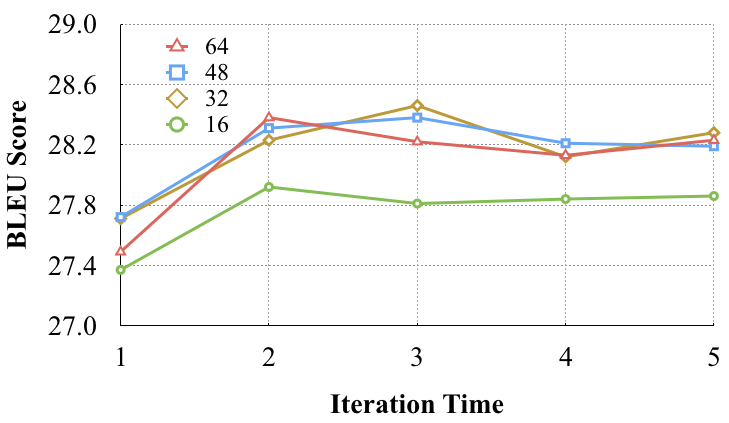}
    \caption{\label{fig: it} The comparison of the GTR with different number of capsules at different iteration times on the EN$\rightarrow$DE task.}
\end{figure}
\subsection{Ablation Study} 
To further show the effectiveness and consumption of each module in our \method model, we make ablation study in this section.
Specifically, we investigate how the \textit{capsule network}, \textit{aggregate structure} and \textit{gating mechanism} affect the performance of the global representation. 

The results are shown in Table \ref{tab: ablation}. Specifically, without the capsule network, the performance decreases 0.7 BLEU , which means extracting features from local representations iteratively could reduce redundant information and noisy. This step determines the quality of global representation directly. Then, aggregating multi-layers' representations attains 0.61 BLEU improvement. The different aspects of information from each layer is an excellent complement for generating the global representation.  
Without the gating mechanism, the performance decreases 0.24 BLEU score which shows the context gating mechanism is important to control the proportion of using the global representation in each decoding step. While the \method model will take more time, we think it is worthwhile to improve generation quality by reducing a bit of efficiency in most scenario.

\begin{table}[t]\footnotesize
    \centering
    \begin{tabular}{l||ccc}
    \toprule
    \multirow{2}{*}{Model}&\multicolumn{3}{c}{Precision}\\
    \cline{2-4}
    &Top-200&Top-500&Top-1000 \\
    
    \hline
    \textit{Last}&43\% &52\%&64\% \\
    \hline
    \textit{Average}&49\%& 57\%&69\%\\
    \hline
    \method&63\%&74\%&81\%\\
    \bottomrule
    \end{tabular}
    \caption{The precision from the bag-of-words predictor based on \method, last encoder state (\textit{Last}) and averaging all local states (\textit{Average}) on the EN$\rightarrow$DE task.}
    \label{tab: pooling}
\end{table}

\begin{figure}[t]
    \centering
    \includegraphics[scale = 0.25]{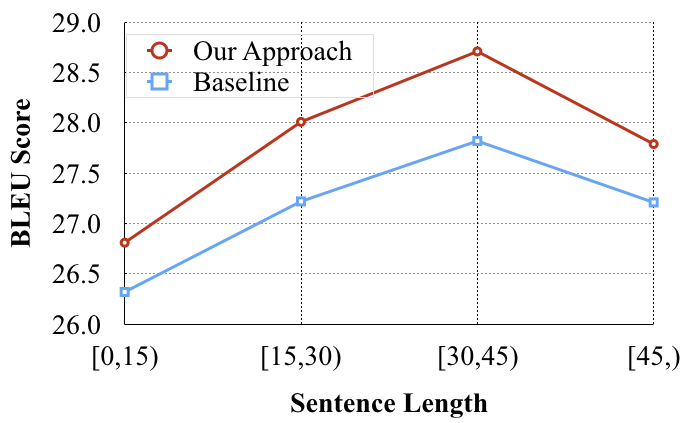}
    \caption{\label{fig: sl} The comparison of the GTR with different number of capsules at different iteration times on the EN$\rightarrow$DE task.}
\end{figure}

\subsection{Effectiveness on Different Model Settings}
We also experiment the \method model with \textit{big} setting on the EN$\rightarrow$DE task.
The \textit{big} model is far larger than above \textit{base} model and get the state-of-the-art performance in previous work~\cite{vaswani2017attention}.

The results are shown in Table \ref{tab: bigsize}, Transformer-Big outperforms Transformer-Base, while the GRET-Big improves 0.86 BLEU score comparing with the Transformer-Big. 
It is worth to mention that our model with base setting could achieve a similar performance to the Transformer-Big, which reduces parameters by almost 75\% (68.3M VS. 249M) and inference time by almost 27\% (0.81x VS. 0.56x).

\begin{figure*}[t]
    \centering
    \includegraphics[scale = 0.9]{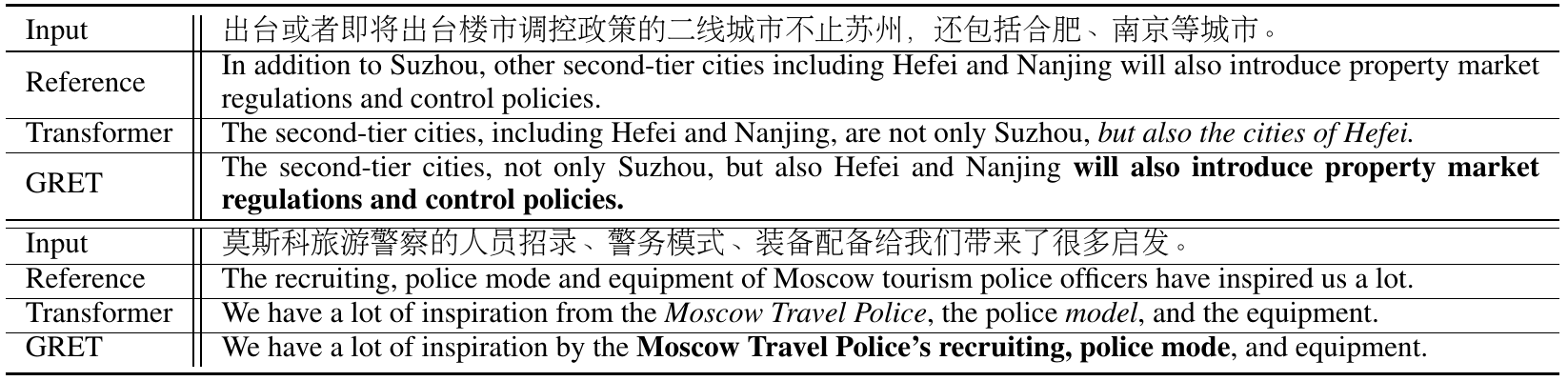}
    \caption{\label{translation samples}Translation cases from Transformer and our \method model on the ZH$\rightarrow$EN task.}
\end{figure*}
\subsection{Analysis of the Capsule}
The number of capsules and the iteration time from dynamic routing algorithm may affect the performance of the proposed model. We evaluate the \method model with different number of capsules at different iteration times on the EN$\rightarrow$DE task. The results are shown in Figure \ref{fig: it}.

We can get two empirical conclusions in this experiment. First, the first three iterations can significantly improve the performance, while the results of more iterations (4 and 5) tend to stabilize. Second, the increase of capsule number (48 and 64) doesn't get a further gain. We think the reason is that most sentences are shorter than 50, just the suitable amount of capsules can extract enough features.

\subsection{Probing Experiment}
What does the global representation learn is an interesting question. Following \newcite{weng2017neural}, we do a probing experiment here. We train a \textit{bag-of-words predictor} by maximizing $ P(\textbf{y}_{bow}|\textbf{s}^{M})$, where $\textbf{y}_{bow}$ is an unordered set containing all words in the output sentence. The structure of the predictor is a simple feed-forward network which maps the global state to the target word embedding matrix.

Then, we compare the precision of target words in the top-\textit{K} words which are chosen through the predicted probability distribution\footnote{Experiment details are shown in \newcite{weng2017neural}.}. The results are shown in Table \ref{tab: pooling}, the global state from \method can get higher precision in all conditions, which shows that the proposed method can obtain more information about the output sentence and partial answers why the \method model could improve the generation quality.

\subsection{Analysis of Sentence Length}
To see the effectiveness of the global representation, we group the EN$\rightarrow$DE test set by the length of the input sentences to re-evaluate the models. The set is divided into 4 sets. Figure \ref{fig: sl} shows the results. 
We find that our model outperforms the baseline in all categories, especially in the longer sentences, which shows that fusing the global representation may help the generation of longer sentences by providing more complete information.

\subsection{Case Study}
We show two real-cases on the ZH$\rightarrow$EN task to see the difference between the baseline and our model. These cases are shown in Figure \ref{translation samples}. The “Source” indicates the source sentence and the “Reference” indicates the human translation. The \textbf{bold font} indicates improvements of our model; and the \textit{italic font} indicates translation errors.

Each output from \method is decided by previous state and the global representation. So, it can avoid some common translation errors like over/under translation, caused by the strong language model of the decoder which ignores some translation information. 
For example, the over translation of ``\textit{the cities of Hefei}'' in case 1 is corrected by the \method model.  Furthermore, providing global information can avoid current state only focuses on the word-to-word mapping. In case 2, the vanilla Transformer translates the ``Moscow Travel Police'' according to the source input ``mosike lvyou jingcha''', but omits the words ``de renyuan zhaolu'', which leads it fails to translate the target word ``\textit{recruiting}''.

\section{Related Work}
Several work also try to generate global representation. In machine translation, \newcite{lin2018de} propose a deconvolutional method to obtain global information to guide the translation process in RNN-based model. However, the limitation of CNN can not model the global information well and there methods can not employ on the Transformer.
In text summarization, \newcite{Chen2018} also propose to incorporate global information in RNN-based model to reduce repetition. They use an additional RNN to model the global representation, which is time-consuming and can not get the long-dependence relationship, which hinders the effectiveness of the global representation.

\newcite{zhang2018sentence} propose a sentence-state LSTM for text representation. Our method shows an alternative way of obtaining the representation, on the implementation of the Transformer. 

Many previous researches notice the importance of the representations generated by the encoder and focus on making full use of them. \newcite{wang2018towards} propose to use Capsule network to generate hidden states directly, which inspire us to use capsules with dynamic routing algorithm to extract specific and suitable features from these hidden states.
\newcite{wang2018multi,dou2018exploiting} propose to utilize the hidden states from multiple layers which contain different aspects of information to model more complete representations, which inspires us to use the states in multiple layers to enhance the global representation.

\section{Conclusion}
In this paper, we address the problem that Transformer doesn't model global contextual information which will decrease generation quality. Then, we propose a novel \method model to generate an external state by the encoder containing global information and fuse it into the decoder dynamically. Our approach solves the both issues of how to model and how to use the global contextual information.
We compare the proposed \method with the state-of-the-art Transformer model. Experimental results on four translation tasks and one text summarization task demonstrate the effectiveness of the approach. In the future, we will do more analysis and combine it with the methods about enhancing local representations to further improve generation performance.

\section*{Acknowledgements}
We would like to thank the reviewers for their insightful comments. Shujian Huang is the corresponding author. This work is supported by the National Key R\&D Program of China (No. 2019QY1806), the National Science Foundation of China (No. 61672277), the Jiangsu Provincial Research Foundation for Basic Research (No. BK20170074).
\bibliography{aaai20}
\bibliographystyle{aaai}

\end{document}